\documentclass[final,5p,times,twocolumn]{elsarticle}
\usepackage{siunitx}
\usepackage{multirow,subfigure}
\usepackage{setspace,longtable}
\usepackage{array}

\newcolumntype{L}[1]{>{\raggedright\let\newline\\\arraybackslash\hspace{0pt}}m{#1}}
\newcolumntype{C}[1]{>{\centering\let\newline\\\arraybackslash\hspace{0pt}}m{#1}}
\newcolumntype{R}[1]{>{\raggedleft\let\newline\\\arraybackslash\hspace{0pt}}m{#1}}

\begin{document}
\title{
How word semantics and phonology affect handwriting of Alzheimer's patients: a machine learning based analysis
%Cognitive impairment diagnosis through handwritten words analysis: from feature selection to classification}
%
%Phonological and semantic handwriting information for Alzheimer's recognition/prediction: a machine learning based analysis}
%
%A word analysis for cognitive impairment diagnosis: from handwriting to classification
%Cognitive impairment diagnosis through handwriting  analysis
%
% Vecchio titolo (WIVACE): 
% How Word Choice Affects Cognitive Impairment Detection by Handwriting Analysis: A Preliminary Study
%Proposta Francesco
% An experimental study on how word meanings affect cognitive impairment diagnosis through handwriting analisys 
%
}
\author[inst1,inst2]{Nicole D. Cilia}
\ead{nicoledalia.cilia@unikore.it}
\author[inst3]{Claudio De Stefano}
\ead{destefano@unicas.it}
\author[inst3,cor1]{Francesco Fontanella}
\ead{fontanella@unicas.it}
\author[inst1]{Sabato Marco Siniscalchi}
\ead{marco.siniscalchi@unikore.it}

\affiliation[inst1]{organization={Department of Computer Engineering, University of Enna "Kore"},%Department and Organization
%            %addressline={Via Giovanni Paolo II 132}, 
%            %city={Fisciano (SA)},
%            %postcode={84084}, 
 country={Italy}
}

% Department of Computer Engineering, University of Enna "Kore", Cittadella Universitaria, Enna, Italy

\affiliation[inst2]{organization={Institute for Computing and Information Sciences, Radboud University Nijmegen},%Department and Organization
            %addressline={Via Giovanni Paolo II 132}, 
            %city={Fisciano (SA)},
            %postcode={84084}, 
            country={The Netherlands}
}
\affiliation[inst3]{organization={Department of Electrical and Information Engineering  Mathematics, University of Cassino and Southern Lazio},%Department and Organization
%            %addressline={Via Giovanni Paolo II 132}, 
%            %city={Fisciano (SA)},
%            %postcode={84084}, 
              country={Italy}
}            
\cortext[cor1]{Corresponding author}

%
%\titlerunning{}

\bibliographystyle{elsarticle-num}

% Include the name of the author that should appear in the running header
%\runningauthor{}%{Cilia et al.}
%\runningauthor{Cilia et al.}

\begin{abstract}
Using kinematic properties of handwriting to support the diagnosis of neurodegenerative disease is a real challenge: non-invasive detection techniques combined with machine learning approaches promise big steps forward in this research field. In literature, the tasks proposed focused on different cognitive skills to elicitate handwriting movements. In particular, the meaning and phonology of words to copy can compromise writing fluency. In this paper, we investigated how  word semantics and phonology affect the handwriting of people affected by Alzheimer's disease.  To this aim,  we used the data from six handwriting tasks, each requiring copying a word belonging to  one of the following categories: regular (have a predictable phoneme-grapheme correspondence, e.g., cat), non-regular (have atypical phoneme-grapheme correspondence, e.g., laugh), and non-word (non-meaningful pronounceable letter strings
that conform to phoneme-grapheme conversion rules). 
We analyzed the data using a machine learning approach by implementing four well-known and widely-used classifiers and feature selection.
The experimental results showed that the feature selection allowed us to derive a different set of highly distinctive features for each word type.
Furthermore, non-regular words needed, on average, more features but achieved excellent classification performance: the best result was obtained on a non-regular,  reaching an accuracy close to 90\%.
\end{abstract}

% Please include a maximum of seven keywords
%\keywords{Word Analysis, Handwriting, Classification Algorithm, Feature Selection, Alzheimer's Disease} %Cognitive Impairment.}

\maketitle
%%%%%%%%%%%%%%%%%%%%%%%%%%%%%%%%%%%%%%%%%%

\section{Introduction}
\label{Sec:Intro}
In recent decades, human movement motor activities have been investigated with increasing frequency with the aim of discovering the underlying cognitive processes. Researchers have found that cognitive processing and the brain motor system are not functionally independent: an individual movement is the end result of a cognitive process. They also found that the relationship between the two systems is much more complex than previously imagined \cite{Shapiro2010}.
Furthermore, more recent findings have shown that alterations in motor activities can be a prodromal sign  of neurodegenerative diseases. For example, people affected by Alzheimer's Disease (AD) exhibit alterations in spatial organization and poor control of fine movements. %This implies that, at least in principle, some diagnostic signs of AD should be detectable by motor tasks. 
In this context, the analysis of the alterations in handwriting can be very useful since handwriting is the result of complex interactions between bio-mechanical parts (arm, wrist, hand, etc.) and brain areas devoted to the control and memorization of the elementary motor sequences used to produce handwritten traces \cite{Impedovo19}. For example, in the clinical course of AD, dysgraphia occurs both during the initial phase and in the progression of the disease \cite{Impedovo18}. It follows that handwriting alterations can be used as further evidence of the onset of AD, helping physicians make an early diagnosis, which remains challenging. 
It is worth noting that early diagnosis is particularly important for AD: the currently available treatments are much more effective in slowing the course of this incurable disease when started early. Even more importantly, also disease-modifying pharmacological treatments that will be available in the near future have shown to be effective only if started in the early stages of AD \cite{Nicoll19}.

Until a few years ago, most of the studies analyzing the effect of AD on handwriting were conducted by physicians and psychologists. They typically based their analysis on data collected from a few dozen participants while performing a few handwriting tasks \cite{Drot16}. Furthermore, those studies typically used statistics-based approaches, e.g., the Pearson correlation coefficient, focused on the relationship between the disease and each used variable (features), overlooking the complex and multiple interactions between them.   
On the other hand, in the last few years, researchers have been interested in applying machine learning (ML) techniques to identify effectively the handwriting peculiarities of AD people. This interest has also been favored by the availability of larger datasets and larger sets of ad-hoc features proposed by experts working on handwriting analysis \cite{Cilia18b}.
%TODO: metter anche IEEEFEATURES : andrebbero speigate le differenze con il lavoro attuale dicendo ceh lì abbiam fatto unanalisis delle features e vistose effettivamente le lafeatuer selection consente di migliaore le preformance 
In general, the benefit derived from ML techniques is twofold. On the one hand, state-of-the-art  ML algorithms typically outperform statistics-based approaches since they are able to better exploit the complex information contained in the data by analyzing  the whole set of features used, not just one feature at a time. On the other hand, analyzing the models built by machine learning algorithms allows users to identify the subset of relevant features for the prediction task, i.e., those actually useful to distinguish the samples belonging to different classes. %marco propone di togliere of the problem at hand.
For example, feature selection tools can be used to improve classification performance and deepen the analysis of the features that better discriminate the classes of the problem at hand independently from the algorithm used. Then these analyses allow effective extraction of knowledge from the data collected, which can be very useful for domain experts. In the case of handwriting analysis to support the diagnosis of AD, since each feature is related to different aspects of the cognitive skills needed to perform a given handwriting task \cite{Impedovo19}, a machine learning based feature analysis may provide physicians with further hints about the brain damages caused by AD.

Studies based on handwriting analysis  have confirmed that 
%machine learning can be a valid tool to identify the handwriting of people affected by AD and that 
different handwriting tasks are characterized by different subsets of relevant features \cite{Cilia22,Cilia21b}. In \cite{Cilia22}, we presented some benchmark results achieved on a dataset containing handwriting from 180 participants while performing copy, drawing, and memory tasks; whereas in \cite{Cilia21b}, we used feature selection to characterize the best set of discriminating features for each task category.

Among the handwriting tasks, those requiring the copy of words aim to test people's abilities in repeating complex graphic gestures. These tasks involve semantic memory and consist of two steps: recognizing the word to be copied and recovering the memory plans needed to write that word.
Semantic memory refers to our general world knowledge encompassing memory for concepts, facts, and the meanings of words and other symbolic units that constitute formal communication systems such as language or math \cite{McRae13}. Modern psychology views semantic memory as profoundly rooted in sensorimotor experience, abstracted across many episodic memories to highlight the stable characteristics and mute the idiosyncratic ones. Recent neuroscience research has focused on how the brain creates semantic memories and which brain regions are responsible for storing and retrieving semantic knowledge \cite{Kumar2021}. % Noppeney09 tolto

According to phonology, the words in a language can be regular and irregular. Regular words have a predictable phoneme-grapheme correspondence (e.g., cat); whereas irregular words have atypical phoneme-grapheme correspondences (e.g., laugh). Semantic also introduces Non-words or pseudo-words, which are non-meaningful pronounceable letter strings that conform to phoneme-grapheme conversion rules \cite{Luzzatti03}. It has been found that each of these word categories can provide specific information about people's cognitive skills \cite{Vessio19}, and % TOLTO: Plat93
%
%\textbf{In particular} \\
%nella letteratura si è visto che la semnatica e la tipologia di parole (SEMANTIC MEMORY) GIOCA UN RUOLO IMPORTANTE PER CAPIRE TANTI ASPETTI\\
%SEMANTIC MEMORY \\
%GLI APSETTI COGNITIVI SONO AMPI E LE DIVERSE PAROLE NE METTONO IN LUCE ASPETTI DIVERSI \\
%
%In the literature can be found 
several studies have investigated how neurodegenerative diseases affect the handwriting of the words belonging to the three above-mentioned word categories \cite{Kolinsky22}. %From the analysis if this 
%PAVED THE WAY TO THE USE OF mER 
%
%pave the way to verificare aspetti cognitivi legati al significato \\
%RELATED WORK SOTTO DIRE L'IMOPRTANZA DI QUWESTE COSE E PRENDERE UN PEZZO PER FAR CAPIRE L'IMOPORTANZA DELLO STUDIO  PER VERIFICARE GLI ASPETTI COGNITIVI DELLE 
%
However, most of those studies did not analyze measurable quantities, such as duration, pressure, and velocity,
% NEI LAVORI PRECEDNTI HANNO MOSTRATO CHE POSSONO ESSERE EFFICACI NEL DISTINGUERE 
but they were mainly concerned with mistakes made in terms of substitutions, or inversion of letters. 
%  \cite{Barca02}.

%In fact, these types of words have been extensively studied in the literature. However, most of the previous studies focused on the handwriting of words belonging to these categories did not analyze measurable quantities such as duration, pressure, velocity, etc. but investigated the mistakes made in terms of substitutions or inversions of letters. 
The lack of studies  based on  the analysis of measurable quantities and the use of machine learning techniques was mainly caused by the lack of an adequate dataset (especially regarding the number of  participants) containing tasks involving the copy of words belonging to the above-mentioned categories. More recently, the dataset presented in \cite{Cilia22} has allowed us to investigate further the role of different tasks in diagnosing AD. 

%\textbf{This dataset also includes data from the three categories of words mentioned above}\\\\
%DA CAPIRE COME COLLEGARE CON SOPRA: \\
%More recently,  
%SCOPO\\
%OPPURE\\
%Starting from the above considerations, 
In \cite{Cilia19c}, we presented a preliminary study in which we investigated how word semantics affects the performance  of a ML system in detecting the handwriting of people affected by AD. 
The rationale of our approach was to use the kinematic and pressure properties of handwriting by using some standard features proposed in the literature to investigate to what extent word semantics affect the handwriting of people affected by AD.
The results presented in \cite{Cilia19c} were obtained on a part of the dataset mentioned above \cite{Cilia22}, and  by using two classification algorithms, namely decision trees and random forests.
%Cognitive Impairment Detection by Handwriting Analysis: A preliminary results on obtained   to compare , with the goal of characterizing words and non-words handwritten by AD people.

%in which we used a part of the this dataset to compare the performance of two classification algorithms (decision trees and random forests) in identifying words and non-words handwritten by people affected by AD. Those results were encouraging and paved the way to further researches. 

The work presented here extends that introduced  in \cite{Cilia19c} along three directions. Starting from a larger number of people recruited for the experiment (99 in the previous study), we included an analysis of the non-regular words. Furthermore, we used more classification schemes and a feature selection technique. Finally, to validate  our results, we compared them with those achieved by the state-of-the-art in handwriting analysis and brain imaging. %Specifically, we added Support Vector Machine and Multi Layer Perceptron as classification schemes and we employed a Recursive Feature Elimination with cross validation for reduce the feature space, in order to identify the most significant features for each different type of word.
%
%(can you be more specific about the feature selection scheme, since you mentioned that in the abstract as a main contribution? How the feature selection technique helped you gain more insights about your problem?)
%
%Specifically, we did .... what??? The analysis allowed us to better understand how AD affects the cognitive skills involved in handwriting copy that specific category of words.
%In this paper, we present the results of those researches, also including the analysis of non regular words. 
%This further analysis allowed us to know how AD affects the cognitive skills involved in copying this category of  words.
We used the extended work presented here to answer the following questions about the semantic and phonological difference between regular, non-regular, and non-words: Is there a task that outperforms the others in predicting the person's cognitive status?
Among the three categories of words considered, is there one that outperforms the others?
Among the features we extracted, are there some that are features that are more relevant than others for all tasks? Or is there a specific subset of features for each word category?

The experimental results confirmed that the feature selection allowed us to improve performance, selecting a different set of features for each type of word, confirming that exploiting different semantic structures can be a good tool for physicians to investigate AD through handwriting analysis.

%ED HANNO FORNITO SPONTI DI SICURO INTRESSE PER IM MEDICI. RISUKATO DEL CONFRONTO 
% RISULTATI SPERIMENTALI HANNO DATO SPUNTI INTERESSANTI SULLE FEATURES ECC. 
% E CHE OGNUNO DI QST TASK È CARATTERIZZATO DA FEATURES DIVERSE

The rest of the paper is organized as follows: Section \ref{Sec:rel} discusses the related work. Section \ref{Sec:Prot} 
describes how the data have been collected and the features extracted. Section \ref{Sec:exp} describes the experiments we have performed and reports the results we have obtained.  
Concluding remarks % and outline of possible future directions of our work 
are eventually left to Section \ref{sec:concl}.

\section{Related Work}
\label{Sec:rel}
% cCOME DETTO IN PRECECENZA LA MAGGOR PARTE  DEI LAVORI SU PAROLO E NON-PAROLE SONO STATE FATTE DA CLIMICI (METTERE ALLA FINE?) PERTANTO QUESTO È IL PRIMO STUDIO CHE 
%COM POSSIAMO VEDERE IN QUEST AMBITO TUTTI GLI STUDUN SONO IN MBITI CLINICOA BMANCA UNA SUTDIUO IN CUI MACHIME LEARNING È USATO PER CAPIER QUALID FEATUERES E 
% c'è l'esigenza di usare  ML per approfondire il ruolo delle featuers e delle parole  del ruolo giocato dalla semnatica  nell predizione d del ad per mezzo del hebadwrting 

%che ruoilo giioca il significato?
%isamoa ML per semnatica del 
%CAPIRE L'IMPORNTANZA DI QUESTE COSE 

As mentioned in \ref{Sec:Intro}, copying words requires first recognizing the word to be copied and then recovering the memory plans needed to write that word. 
It is well known that numerous factors affect the recognition of written words. Among these, the frequency of use is relevant: more frequent words are easier to recognize. Another critical factor is word length: shorter words are easier to recognize \cite{Weekes97}.
Many other factors have specific importance, for example, the age of acquisition (the time the word was learned) or the spelling proximity (each word has a certain degree of similarity with few or many other words in the lexicon). In \cite{Barca02}, the authors investigated the impact of these (and other) variables on reading words in adult readers. 
In addition to words with a lexical value, i.e., having precise semantics, tasks can also consist of words with no lexical value (non-words in the following). Execution in non-word reading is informative of the ability to decode stimuli using the grapheme-phoneme conversion rules without a lexicon contribution. Also, in this case, the proposed stimulus (word) length plays an important role \cite{Zoccolotti05}.

The importance of word semantics has also been investigated in children in the  developmental age. In the work presented in \cite{Hasen21}, the authors investigated the interaction between orthographic and morphological processes in meaning activation during reading. To this aim, they asked 80 primary school children to make word category decisions. Results revealed that words were harder to reject as category members when the embedded stem was category-congruent. In \cite{Richlan20}, the author  reviewed the literature on the functional neuroanatomy of developmental dyslexia across languages and writing systems. The review included comparisons of alphabetic languages differing in orthographic depth as well as comparisons across alphabetic, syllabic, and logographic writing systems. It also synthesized the evidence for universal and language-specific effects on dyslexic functional brain activation abnormalities during reading and reading-related tasks. 
%, for example \cite{Burani02, Ziegler03}, 
%BURANI: How early does morpholexical reading develop in readers of a shallow orthography?
%ZIEGLER: Developmental dyslexia in different languages: Language-specific or universal?

In the study of neurodegenerative diseases, the choice of different types of words more useful to bring out the symptoms of AD has also assumed importance. Handwriting analysis through machine learning techniques highlighted the diagnostic power of using words with full meaning (regular and non-regular) and non-words.
For example, in \cite{Vessio19}, the author presented a literature review of the research investigating the nature of writing impairment associated with AD. He found that in most studies, words are usually categorized as regular, irregular, and non-words. 
Orthographically regular words have a predictable phoneme-grapheme correspondence (e.g., cat), whereas irregular words have atypical phoneme-grapheme correspondences (e.g., laugh). Non-words or pseudo-words are non-meaningful pronounceable letter strings that conform to phoneme-grapheme conversion rules and are often used to assess phonological spelling. In \cite{Plat93}, the authors proposed a writing test from dictation to 22 patients twice, with an interval of 9-12 months between the tests. They found that agraphic impairment evolved through three phases in patients with AD. The first one is a phase of mild impairment (with a few possible phonologically plausible errors). In the second phase, non-phonological spelling errors predominate, phonologically plausible errors are fewer, and the errors mostly involve irregular words and non-words. The study of \cite{Pestell00} investigated handwriting performance on a written and oral spelling task. The authors selected thirty-two words from the English language: twelve regular words, twelve irregular words, and eight non-words. The study aimed to find logical patterns in spelling deterioration with disease progression. The results suggested that spelling in AD individuals was impaired relative to the healthy control group (HC). Finally, \cite{Luzzatti03} used a written spelling test made up of regular words, non-words, and words with unpredictable orthography. The study aimed to test cognitive deterioration from mild to moderate AD. The authors found little correlation between dysgraphia and dementia severity.

From the brief literature review outlined above, we can argue that most of the previous studies focusing on the handwriting of words belonging to these categories did not analyze measurable quantities such as duration, pressure, velocity, etc., but investigated the mistakes made in terms of substitutions or inversions of letters. 
To the best of our knowledge, this is the first study in which ML techniques are used to investigate how word semantics affect the handwriting of people affected by AD.

\begin{figure*}[t]
	\centering
	\subfigure[Regular words]{\includegraphics[scale=0.6] {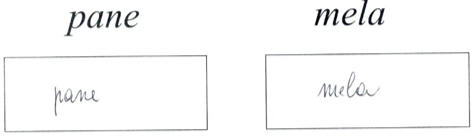}} %0.8
	\subfigure[Non-Regular words]{\includegraphics[scale=0.6]{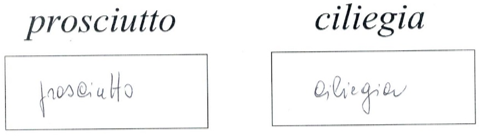}}
	\subfigure[Non-words]{\includegraphics[scale=0.6]{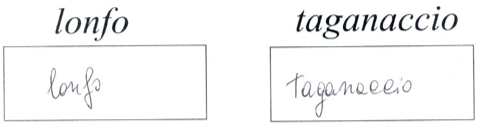}}
	\caption{Examples of the six tasks performed by a participant.}
	\label{Fig:Fig1}
\end{figure*}

\section{Data collection and feature extraction}
\label{Sec:Prot}
In the present work, participants were recruited with the support of the geriatric ward, Alzheimer unit, of the ”Federico II” hospital in Naples. We considered clinical tests (such as PET, TAC, and enzymatic analyses) and standard cognitive tests (such as MMSE) for the recruitment criteria. For both patients and the control group, we checked whether they were on therapy or not, excluding those who used psychotropic drugs or any other drug that could influence their cognitive abilities.
As previously mentioned, the aim of the protocol was to record the dynamics of handwriting in order to investigate whether there are specific features that allow us to distinguish people affected by AD from healthy ones. 
In the present study, we focus on the frequency of use, length, and lexicality of the selected words. The six tasks considered in the study, namely two tasks of Regular Word (RW), two tasks of non-regular Word (NRW), and two tasks of Non-Word (NW), required people to write each word in the appropriate box. As suggested in \cite{Luzzatti03} and \cite{Platel93}, we chose the following words of the Italian language:
\begin{itemize}
	\item[--] RW: "pane" and "mela" (bread and apple in English); 
	\item[--] NRW: "prosciutto" and "ciliegia" (ham and cherry);
	\item[--] NW: "taganaccio" and "lonfo" (nonsense words)
\end{itemize}
RW words represent stimuli with lexical value, in which the age of word acquisition and length are important. On the other hand,  NW words  highlight the ability to decode stimuli using the grapheme-phoneme conversion rules in the absence of a lexicon contribution. Finally, the  stimuli produced by NRW words spur the ability to copy a word representing a stimulus that does not have a perfect match between phoneme and grapheme. Although it may seem that in the Italian language there is almost always full correspondence between grapheme and phoneme, this is not always true. For example, the letter "i" represented by a single grapheme behaves in three different ways: i) it behaves as a vowel phoneme; ii) it behaves as a semiconsonant; iii) it can become a prosthetic "i", that is, it acts as a prosthesis and serves to indicate the palatalization of the previous sound.
\begin{table*}[t!]
	\centering
%{\color{blue} 
\caption{Feature list. Feature types are: dynamic (D), static (S), and personal (P).}
    \scalebox{0.7}{
	\begin{tabular}{clp{9cm}c}
{\# }         & {Name}    & {Description}    & {Type}		 \\ 
\hline
1	&	Duration	&	Time interval between the first and the last points in a stroke	&	D	\\
2	&	Start Vertical Position	&	Vertical start position relative to the lower edge of the active digitizer area	&	S	\\
3	&	Vertical Size	&	Difference between the highest and lowest $y$ coordinates of the stroke	&	S	\\
4	&	Peak vertical velocity	&	Maximum value of vertical velocity among the points of the stroke	&	D	\\
5	&	Peak vertical acceleration	&	Maximum value of vertical acceleration	among the points of the stroke &	D	\\
6	&	Start horizontal position	&	Horizontal start position relative to the lower edge of the active tablet area	&	S	\\
7	&	Horizontal size	&	Difference between the highest (rightmost) and lowest (leftmost) $x$ coordinates of the stroke	&	S	\\
8	&	Straightness error	&	It is calculated by estimating the length of the straight line, fitting the straight line, estimating the (perpendicular) distances of each point to the fitted line, estimating the standard deviation of the distances and dividing it by the length of the line between beginning and end &	D	\\
9	&	Slant	&	Direction from the beginning point to the endpoint  of the stroke, in radiant	&	S	\\
10	&	Loop Surface	&	Area of the loop enclosed by the previous and the present stroke	&	S	\\
11	&	Relative initial slant	&	Departure of the direction during the first 80 ms to the slant of the entire stroke.	&	D	\\
12	&	Relative time to peak
vertical velocity	&	Ratio of the time duration at which the maximum peak velocity occurs (from the start time) to the total duration	&	D	\\
13	&	Absolute size	&	Calculated from the vertical and horizontal sizes	&	S	\\
14	&	Average absolute velocity	&	Average absolute velocity computed across all the samples of the stroke	&	D	\\
15	&	Road length	&	length of a stroke from beginning to end, dimensionless	&	S	\\
16	&	Absolute y jerk	&	The root mean square (RMS) value of the absolute jerk along the vertical direction, across all points of the stroke	&	D	\\
17	&	Normalized y jerk	&	Dimensionless as it is normalized for stroke duration and size	&	D	\\
18	&	Absolute jerk	&	The Root Mean Square (RMS) value of the absolute jerk across all points of the stroke	&	D	\\
19	&	Normalized jerk	&	Dimensionless as it is normalized for stroke duration and size	&	D	\\
20	&	Number of peak acceleration points	&	Number of acceleration peaks both up-going and down-going in the stroke	&	S	\\
21	&	Pen pressure	&	Average pen pressure computed over the points of the  stroke	&	D	\\
22	&	\#strokes	&	Total number of strokes of the task	&	S	\\
23	&	Sex	&	Participant's gender	&	P	\\
24	&	Age	&	Participant's age	&	P	\\
25	&	Work	&	Type of work of the participant (intellectual or manual)	&	P	\\
26	&	Instruction	&	Participant's education level, expressed in years 	&	P \\
\end{tabular}
}
\label{tab:feat}
%}
\end{table*}

The aim of these six tasks is to compare the features extracted from handwriting movements of these six different types of words. We chose the six tasks according to the following criteria:
\begin{itemize}
	\item[(i)] The copy tasks of NW allow us to compare the writing variations with regard to the reorganization of a specific motor plan, present or absent in the subject’s memory.
	\item[(ii)] Tasks need to involve different graphic arrangements, e.g., words with ascenders and/or descendants, allowing us to test fine motor control capabilities. Indeed, the (RWs) have different descender (the "p" of the first word) and ascender traits (the "l" in the second word). The NWs propose the same structure: the first word has descender traits (the "g" in "taganaccio") and the ascender traits in the second one (the "l" and "f" in "lonfo"). Note that the NWs must be built following the synthetic rules of language chosen.
	\item[(iii)] Tasks need to involve different pen-ups to analyze air movements, which are typically altered in AD patients.
	\item[(iv)] We have chosen to present the tasks by asking the people to copy each word in the appropriate box. Indeed, according to the literature, the box allows the assessment of the spatial organization skills of the patient.	
\end{itemize}

Fig. \ref{Fig:Fig1} shows an example of the six tasks performed.   %collectechosen for this study , with different ascender and descender traits, 
To collect the data, we used a graphic tablet to record the movements of the pen used by the examined person in terms of x,y, and z (pressure) coordinates, sampled at a frequency of 200Hz.

\subsection{Feature extraction}
\label{Sec:Feat}
The features extracted from the available raw data, i.e., $(x,y)$ coordinates, pressure, and timestamps, were calculated on the strokes making up the handwritten traits and then averaged over the entire task. 
We extracted both static and dynamic features. The former were computed considering the strokes' shape or position; whereas the second were computed on the velocity profile. Table \ref{tab:feat} shows the list of the extracted features. 
A stroke is defined as the single component making up a handwritten trait, and it is represented by the sequence of points between two consecutive segmentation points. We considered as segmentation points: pen-up, pen-down, and zero-crossing velocity along the y-axis. 
Since many studies in the literature show significant differences in the motor performance in patients between in-air and on-paper traits \cite{Impedovo19b}, each feature was calculated separately for the in-air or on-paper traits. In particular, we extracted three groups of features:
 
\begin{itemize}
\item In-air: the features extracted from the in-air traits (acquired by the system when the pen is lifted from the sheet within the maximum acquisition distance of 3 cm). These movements characterize the planning activity for positioning the pen tip between two successive written traits. Note that in this case the pressure feature (\#21 in \ref{tab:feat}) is not used. These features will be denoted in the following with the letter 'A';
\item On-paper: the features extracted from the written traits (during a pen-down and the successive pen-up). These features will be denoted with the letter 'P';
\item All: in this case, each sample is represented by a feature vector containing both in-air and on-paper features. Note that in this case, the total number of  features is 47 (personal features and pressure are not repeated). The aim is twofold. On one hand, we want to perform a direct comparison between in-air and on-paper features, on the other hand, we want to investigate the interactions between in-air and on-paper features. These features will be denoted as "AL".
\end{itemize}

In order to take into account the differences due to age, education, or work, we have also added the following "personal" features: gender, age, type of work, and level of education.
In summary, we used three groups of features, each represented by six tasks, each containing 180 samples represented  by the features discussed. In particular, we extracted 26 features (see Table \ref{tab:feat}) for the on-paper (P), 25 in-air features (A) because pen pressure is not used, and 44 for the  category All (AL).

\section{Experimental Results}
\label{Sec:exp}
To investigate how word semantics affects the handwriting of people affected by AD and what performance can be achieved by a machine learning based solution on AD detection using handwriting, we implemented the experimental methodology shown in Fig.~\ref{Fig:Fig1}. 
Starting from the raw handwriting data of the six words (Section \ref{Sec:Prot}), we extracted the features detailed in Section \ref{Sec:Feat}. 
Afterwards, we used the following ML algorithms to put forth our binary classification task:
%this data to train  four well-known and widely used classifiers, namely 
Random Forest (RF) \cite{RFcit}, Decision Tree (DT), Support Vector Machine (SVM) \cite{Kee03}, Multilayer Perceptron (MLP) \cite{Rumelhart}. To test the performance achieved by these classifiers we used a ten-fold cross-validation strategy. Then we performed four sets of experiments: 
\begin{itemize}
    \item[(i)] In the first set we tested the performance on every single task (in total six tasks) by separately evaluating the contribution to each task of the P, A, and AL features; 
    \item[(ii)] In the second set, we grouped the six tasks into three groups, each containing two tasks of the same type, namely, regular words (RW), non-regular words (NRW), and non-words (NW). In practice, for each group, each classifier was trained on the samples belonging to the two tasks of a given category. 
    \item[(iii)] In the third set, we evaluated the classification performance as a result of feature selection. As explained in Section \ref{sec:FeatSel} we used the Recursive Feature Elimination (RFE) technique.
    \item[(iv)] To test the effectiveness of the analysis of handwriting as a tool to support the diagnosis of AD, we compared our results with those achieved by  state-of-the-art approaches using two types of data, namely handwriting and brain images.
\end{itemize}
The results obtained from these experiments are detailed in the following subsections. 
\begin{figure*}[t]
	\centering
	\includegraphics[width=\textwidth]{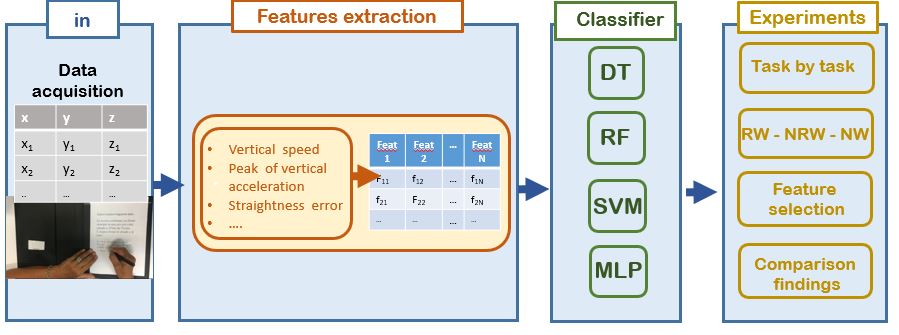}
	\caption{Experimental methodology.}
	\label{Fig1}
\end{figure*}

\subsection{Single task results}
\label{sec:single}
In the first set of experiments, we investigated the performance achieved on each task, by evaluating separately the contribution of the P, A, and AL features. In particular, in this experiment we tried to answer the following question: Is there a task that outperforms the others in the prediction of the cognitive status of the person?

Table \ref{tab:single} shows the results achieved using the four classifiers and the three feature sets considered. From the table \ref{tab:single} we can observe that task \#2 (the regular word "mela") outperformed the other tasks with an accuracy of 82.32\%, using the A features and the DT classifier. Also for the task \#1 we obtained the best classification accuracy (81.21\%) with the features A and with the same classifier DT. This result suggests that regular words are more discriminant than the other categories and that the in-air movements of people affected by AD are more altered while copying regular words.

As for the non-regular words, we can observe that does not seem to be much difference between the two words. The same did not occur for the non-words. Indeed, "taganaccio" allowed us  to obtain a better classification accuracy than "lonfo". That can be explained considering the different lengths of the two words: %The reason can be due to the different length of the words: 
longer words contain more information.
Finally, we can also observe that DT and SVM were the best-performing classifiers.

% \usepackage{multirow}
\begin{table*}[p]
\centering
\caption{Single task accuracies. For each task the best result is in bold.}
	\begin{tabular}{ccccccc}
	\hline
	\multirow{2}{*}{\textbf{Task Type}} & \multirow{2}{*}{\textbf{Task\#}} & \multirow{2}{1cm}{\textbf{Feature Type}} & \multicolumn{4}{c}{\textbf{Accuracy (\%)}}                         \\ \cline{4-7} 
	&                                  &   & \textbf{DT}    & \textbf{RF}    & \textbf{SVM}   & \textbf{MLP}   \\ \hline
	%%%% TASK 1
	%\multirow{6}{*}{RW}                 & \multirow{3}{*}{1}      
	RW & 1 & A & \textbf{81.21} & 77.9           & 74.58          & 70.16          \\ %\cline{3-7}
	&                                  & \textbf{P}                             & 79.55 & 79.55 & 74.58 & 70.71 \\ %\cline{3-7}
	&                                  & AL                                     & 70.16 & 77.34 & 71.27          & 67.4           \\ \cline{2-7}

	%%%% TASK 2
	%& \multirow{3}{*}{2}               
	& 2 & A & \textbf{82.32} & 74.58 & 74.03          & 71.82          \\ % \cline{3-7}
	&                                  & P & 74.58          & 72.92          & 69.61          & 71.82          \\ % \cline{3-7}
	&                                  & AL                                     & 71.27          & 71.27          & 71.82          & 65.74          \\ \hline %\hline
	%%%% TASK 3
	% \multirow{6}{*}{NRW}                & \multirow{3}{*}{3}
	NRW & 3 & A & 76.24 & 79.00          & 76.24          & 70.71          \\ %\cline{3-7}
	&                                  & P & 76.24          & 80.05 & \textbf{80.31} & 75.69          \\ % \cline{3-7}
	&                                  & AL                                & 76.66          & 78.88          & 77.77          & 68.88          \\ \cline{2-7}
	%%%% TASK 4
	% & \multirow{3}{*}{4}
	& 4 & A & 79.55          & 74.58          & 76.79          & 74.03          \\ % \cline{3-7}
	&  & P & 74.03 & 75.13          & 77.34          & 75.13          \\ %\cline{3-7}
	&                                  & AL                                     & 76.79          & 77.34 & \textbf{80.11} & 76.24         \\ \hline 
	%%%% TASK 5
	% \multirow{6}{*}{NW}                 & \multirow{3}{*}{5}
	NW & 5 & A & \textbf{77.9}  & 75.13 & 76.79          & 70.16          \\ %\cline{3-7}
	&                                  & P & 72.37          & 75.13          & 71.27          & 69.06          \\ %\cline{3-7}
	&                                  & AL                                     & 73.48          & 67.4           & 77.34          & 75.13          \\ \cline{2-7}
	%%%% TASK 6
	% & \multirow{3}{*}{6} 
	  & 3 & A & 75.13          & 75.69          & \textbf{79.00} & 71.82          \\ % \cline{3-7}
	&                                  & P & 75.69          & 71.27          & 75.13          & 72.37          \\ %\cline{3-7}
	&                                  & AL                                     & 75.69          & 78.45 & 75.13          & 71.27          \\ \hline 
\end{tabular}

\label{tab:single}
\end{table*}

\subsection{Merged task results}
\label{sec:merge}
In the second set of experiments, we tried to answer the following question: among the three categories of words
considered, is there one that outperforms the others? To answer the question, we grouped the six tasks into three groups (RW, NRW, and NW). In practice, each classifier was trained on the samples belonging to the two tasks of the same category.
Table \ref{tab:merged} summarizes our results.%The results achieved are shown in Table .

\begin{table*}[p]
\centering
\caption{Merged task accuracies. For each group, the best result is in bold.}
%\includegraphics[scale=0.8]{mergedTable.PNG}
% \usepackage{multirow}
	\begin{tabular}{ccccccc}
	\hline
	\multirow{2}{*}{\textbf{Task Type}} & \multirow{2}{*}{\textbf{Task\#}} & \multirow{2}{1cm}{\textbf{Feature Type}} & \multicolumn{4}{c}{\textbf{Accuracy (\%)}}                         \\ \cline{4-7} 
	&                                  &   & \textbf{DT}    & \textbf{RF}    & \textbf{SVM}   & \textbf{MLP}   \\ \hline
	%%%% TASKS 1-2
		RW                 & 1-2             & A & \textbf{85.08} & 80.11          & 75.69          & 72.92          \\  
		&                                  & P                             & 76.1  & 78.17 & 76.4  & 74.92 \\  
		&                                  & AL                                     & 76.4  & 79.05 & 79.05          & 77.28          \\ \hline  
		%%%% TASKS 3-4
		NRW       & 3-4             & A & \textbf{83.97} & 79.83 & 79.83          & 79.28          \\  
		&                                  & P & 78.46          & 77.87          & 79.64          & 76.4           \\  
		&                                  & AL                                     & 78.76          & 83.48          & 82.3           & 77.87          \\ \hline  
		%%%% TASKS 5-6
		NW                & 5-6             & A & 80.93 & 77.9           & 78.72          & 74.03          \\ 
		&                                  & P & \textbf{84.11} & 78.52          & 76.47          & 77.35          \\ 
		&                                  & AL                                     & 82.64          & 79.41 & 79.41 & 76.76        \\ \hline
	\end{tabular}
%\includegraphics[scale=0.3]{tab3.png}
% \usepackage{multirow}
\label{tab:merged}
\end{table*}

Again, regular words achieved the best classification accuracy, with the A features reaching an accuracy of 85.08\%. Most probably, this result is due to the fact that the motor plan for this kind of word is already present in memory. AD harms the brain area containing this memory and causes alterations in the movements needed to write this type of word. Furthermore, in-air movements are more discriminant than on-paper since while performing on-paper movements, people receive the visual feedback of the ink trace he is writing. This allows AD patients to correct the movements altered by brain harms. Clearly, the same cannot occur with the in-air movements because there is no visual feedback for them.
Similar results can also be observed  for non-regular words, which although require the execution of more complex motor plans they are still based on the same memory mechanism.  

%damaged by AD, and cause the    and the brain are  which are damaged and AD  deteriorates in people with AD and remains intact in healthy people where the differences in the motor execution plan are more evident. 
The scenario outlined above does not hold for non-words. Indeed, in this case, the motor plane  is not present in memory but must be generated while the person is writing. From table \ref{tab:merged}, we can see that, in this case,  on-paper features achieved the best performance. Most probably, this is due to the fact that brain areas that allow people to generate the motor planes "on demand" are harmed in people affected by AD. 
It is also worth noticing that the DT achieved the best performance. This result is due to the ability of DTs to use different feature subsets and decision rules at different stages of classification.

Finally, comparing Table \ref{tab:single} (single tasks) and Table \ref{tab:merged} (merged tasks), we can observe that in most cases we achieved better results on the merged tasks. This result suggests that movement anomalies of participants with AD  while writing words belonging to a given category share common characteristics. Consequently, these shared characteristics allow the training phase on the merged datasets (one per category) to be more effective than that obtained on the single data (one per word).      
% dataset + grande 
% However, the differences are not very consistent because in the NW we still used words that follow the syntactic rules of the Italian language. Finally, while in the RW and NRW the most relevant features eere the A ones, for the NW the most performing features were the P ones. In general, the best classifier was J48.

\begin{table*}[t]
	\centering
	\caption{Classification results with feature selection. For each task, the best result is in bold.}
	\begin{tabular}{ccccccc}
		\hline
		\multirow{2}{*}{\textbf{Task Type}} & \multirow{2}{*}{\textbf{Task\#}} & \multirow{2}{1cm}{\textbf{Feature Type}} & \multicolumn{4}{c}{\textbf{Accuracy (\%)}}                         \\ \cline{4-7} 
		&                                  &   & \textbf{DT}    & \textbf{RF}    & \textbf{SVM}   & \textbf{MLP}   \\ \hline
		RW & 1 & A & 78.81          & 80.51          & 77.97          & 80.51          \\  
		&              & P                             & 80.00 & 85.19 & 80.00 & 79.26 \\  
		&                                  & AL                                     & 84.17 & \textbf{86.67} & 80.00          & 80.83          \\ \cline{2-7}
		& 2               & A & 80.85          & \textbf{82.98} & 75.53          & 76.60          \\  
		&                                  & P & 79.41          & 80.15          & 77.21          & 74.26          \\  
		&                                  & AL                                     & 78.68          & 82.35          & 75.00          & 77.21          \\ \hline  
		
		NRW                & 3               & A & \textbf{86.81} & 84.03          & 79.86          & 79.86          \\ 
		&                                  & P & 80.00          & 78.52          & 77.78          & 80.74          \\ 
		&                                  & AL                                     & 82.26          & 84.68          & 77.42          & 75.81          \\ \cline{2-7}
		& 4               & A & 85.71          & 85.71          & 76.19          & 78.57          \\  
		&                                  & P & \textbf{88.97} & 86.76          & 80.15          & 79.41          \\  
		&                                  & AL                                     & 80.95          & 84.13 & 79.37 & 76.98          \\ \hline                     
		NW                 & 5               & A & 81.94          & \textbf{88.19} & 80.56          & 79.86          \\  
		&                                  & P & 77.94          & 87.50          & 71.32          & 72.79          \\  
		&                                  & AL                                     & 75.00          & 84.56          & 78.68          & 75.74          \\  \cline{2-7}
		& 3               & A & 77.61          & 82.84          & 78.36 & 77.61          \\  
		&                                  & P & 86.81          & 85.42          & 77.78          & 80.56          \\ 
		&                                  & AL                                     & 84.56          & \textbf{87.50} & 82.35          & 80.15       \\ \hline    
	\end{tabular}
	\label{tab:selected}
\end{table*}
\subsection{Feature selection}
\label{sec:FeatSel}
In the third set of experiments, we tried  to answer the following question: among the features we extracted, are there some that are more relevant than others? 
To find the features that allow better discrimination between the patients affected by AD and the control group, we used a well-known feature selection technique based on a wrapper evaluation function, named recursive feature elimination (RFE in the following) 
\cite{Guyon03}. 

% Marco suggerisce di inserire questa parte prima nella section 3
RFE performs a greedy search to find the best-performing feature subset based on the backward elimination strategy. Starting from the whole set of available features, the RFE algorithm iteratively creates models. It determines the worst-performing feature at each iteration. Then, it builds the subsequent models with the remaining features until all  features are evaluated. If the data contain $N$ features, RFE evaluates $N^2$ subsets in the worst case. As an evaluation function, we used the accuracy computed using the K-fold cross-validation technique, achieved by using the XGBoost classifier.
%%%%%%%%%%%%%%%%%%%%%%%%%%%%%
The accuracy values of the results are shown in Table \ref{tab:selected}.
From the table we can see that feature selection allowed us to achieve better performances than those achieved using all features (see Table \ref{tab:single}). Looking at the table, it is worth noting that the highest performance increment was achieved on the NRW and NW words. 
% COMMENTATO PER LA VERSIONE FINALE
% Although there is not a valid criterion to explain every trend, we can say that the process we have seen in the two previous experiments is reversed because the tasks that do best are NRW and NW. This could mean that the NW could be very relevant for the diagnosis if supported by the right selection of features.
%
% From the table we can also observe that 
In particular, the fourth task with DT achieved the best performance using the on-paper features, reaching an accuracy of 88.97\%. This result is of particular importance as decision trees have great explanatory power. The second best performance was achieved on task \#5, using the A features with the RF classifier. Finally, the third best performance was achieved on task \#6 with AL features and the same classifier.
%
% COMMENTATO PER LA VERSIONE FINALE
%Looking at Table \ref{tab:selected}, we can also observe that the RF classifier is the best performing  with feature selection. %This result confirm that feature selection allows 

%However, the classification trees are the most performing classifiers with features selection. 
%

\begin{figure*}[p]
	\centering
	\subfigure[Regular words]{\includegraphics[width=\textwidth]{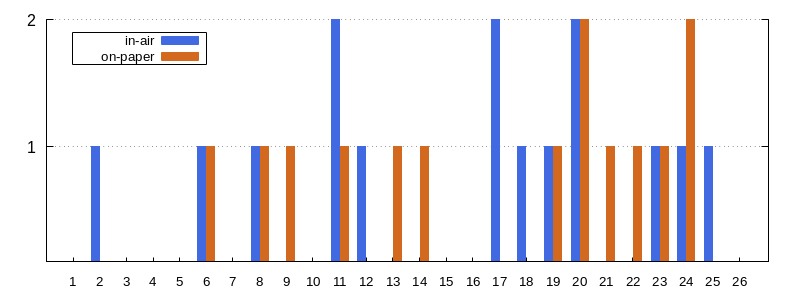}}
	\subfigure[Non-regular words]{\includegraphics[width=\textwidth]{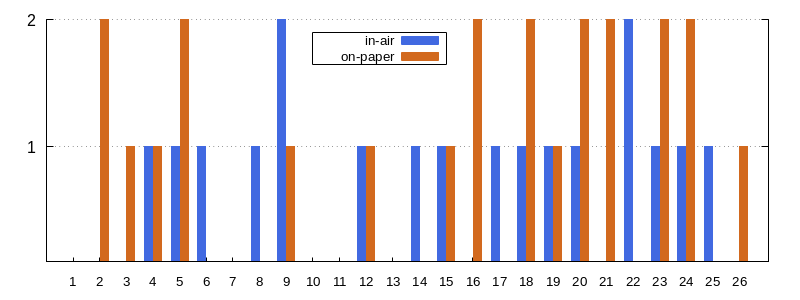}} % era scale=0.6
	\subfigure[Non-words]{\includegraphics[width=\textwidth]{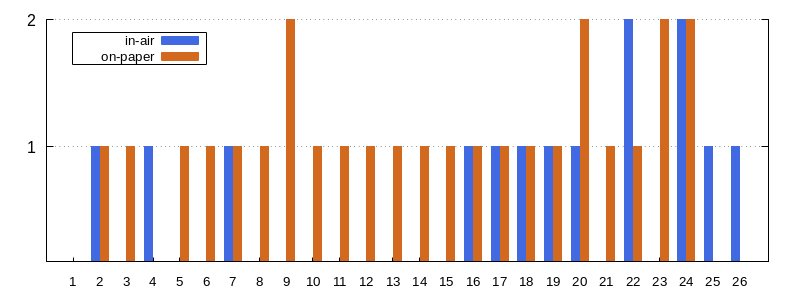}}
	\caption{Occurrence of features for the AL feature set.}
	\label{Fig:histo}
\end{figure*}

To answer the specific question of whether, among the features we extracted, some are more important than others, for each task category, in Figure \ref{Fig:histo} we plotted the histograms  showing how many times RFE selected each feature across the two tasks of a given category. Furthermore, to better investigate the interaction between the in-air and on-paper features, we considered only the AL features, i.e., those obtained merging in a single feature vector in-air and on-paper features (see Section \ref{Sec:Feat}). All histograms are shown in Fig.~\ref{Fig:histo} and commented in the following.

%MEMO:VA DETTO CHE LO STUDIO È STATO FATTO SEAPARATAMENTE:\\
%IN A FIRST STEP WE CONSIDEERED THE FEATURE SELECTED  WHILE USING THE in-air OR ON-PAPER, SEPARATELY. tHESE HISTOGRAMS ARE SHOWN IN TABLE xxx.
%iN A SECOND STEP, WE CONSIDERED THE FEATURES SELECTED from the all set featuresWe remind that in this case, each sample is represented by single feature vector containing both in-air and on-paper features. The aim of this second analisys is \\
%qui lo scopo è quello di capire l'interazione tra queste due diferse tipologie di featues

%\textbf{MEMO 2:\\ 
%NEL DESCRIVERE I GRAFICI SOTTO BISOGNA AVERE CHIARO IN MENTE CHE LE FEATURES SONO STATE SELEZIONATE SEPARATAMENTE		
%} 

%
\subsubsection{Regular words}
For the regular words (top histogram) we can see that nine features were never selected. This fact confirms that the movement anomalies of participants with AD  while writing regular words can be characterized taking into account fewer features than those needed for the other two categories. From the histogram, we can also see that many features were selected only for a single task (ten on-paper and nine in-air). These features represent about 80\% of the 24 features selected at least one time (both in-air or on-paper). 
This result highlights that the movement anomalies of participants with AD while writing the two regular words we considered ("pane" and "mela") are characterized by different features. Actually, this result was unexpected: the two words are both, simple, short, as well as regular. Therefore,  we expected that  the same features were involved in these tasks.
On the contrary, this result confirms that for regular words, the word's specific meaning plays an important role in the cognitive processes involved in the handwriting of this kind of word. The histogram shows that the two words share a few common features: three in-air and two on-paper. However, it is worth noticing that the number of peak acceleration points (feature \#20) is shared by both words and feature types. This result confirms that both in-air and on-paper movements of AD patients are characterized by a different number of acceleration peaks while writing regular words.  
Finally, the histogram also shows that the normalized y jerk (feature \#17) is characteristic only for in-air movements.
%This result will be discussed in the following as comment to the results shown in Table \ref{tab:selected}.
\subsubsection{Non-regular words}
The histogram of the non-regular words (middle plot) shows that the number of non-selected features is much less than that of the regular words (five instead of nine). This result suggests that the movement anomalies of people with AD while writing non-regular words need more features to be characterized with respect to regular words. This result probably depends on the higher complexity of the motor plans of non-regular words with respect to the regular ones. From the histogram, we can also observe that in this case, the features selected for a single task are fourteen and seven, for the in-air and on-paper, respectively (about 68\% of the 31 features selected at least once). 
This value, although lower than that of the regular words (80\%, see above), seems to confirm that also for non-regular words, the specific meaning of the word affects significantly the cognitive processes involved in the handwriting of this kind of words. The remaining ten features are shared between the two tasks (ten on-paper and three in-air).
% This result will be discussed in the following as comment to the results shown in Table \ref{tab:selected}
%
Regarding these "shared features", it is interesting noting that most of them are on-paper. This result highlights the importance of on-paper features in discriminating the handwriting of AD patients while writing non-regular words. This trend will be confirmed and discussed below for non-words.   
\subsubsection{Non-words}
From the bottom histogram we can observe that all features except the duration were selected for the non-regular words. This result confirms the trend already seen above, the higher the complexity of the motor plan needed to write a word, the higher the number of features needed to characterize the handwriting of that word. Furthermore, it is worth noting that the only feature not selected is the duration. This means the handwriting of both healthy people and patients is characterized by similar duration, due to the difficulty of writing words for which they don't have a motor plan since they have never seen them before. 

From the histogram we can also observe that in this case the most selected feature are the on paper ones (26 out of  40 selected features). Most probably, this is due to the fact that the on-paper features  describe movements that are more constrained than the in-air movements because they are constrained by the form of the characters or symbol to be drawn. Then these results confirm that healthy people have a higher ability than AD patients to control the on-paper  movements while writing non-words.          
From the histogram we can also observe that 28 features were selected only for a single task (eighteen  on-paper and ten in-air). These features represent about 80\% of the 34 features selected at least one time. As is the case of regular words, this result confirms that the specific meaning of the word plays an important role in the cognitive processes involved in the handwriting of non-words. Finally, the only feature shared by both words and feature types is age (\#24). This result confirms that the way people (both healthy and affected by AD) write non-words is affected by age, and then it must be taken into account to effectively distinguish healthy people from AD patients.

\subsection{Comparison findings}
Although the main purpose of this study is to investigate how word semantics affects the handwriting of people affected by AD, we also investigated to what extent the three types of words analyzed in this study can be taken into account to implement an effective machine learning based tool to support the diagnosis of AD. To this aim, we compared our performance with those reported in the  literature. In order to deepen as much as possible this comparison, we compared our results with those achieved  on handwriting data \cite{Impedovo18} and on brain images \cite{Ebra20,Goyal2021}. 
It is worth pointing out that this comparison aims to validate the findings detailed in the previous sections, related to investigating how the handwriting of different word categories, each involving different cognitive skills, is affected by AD. Indeed, we think that this validation goes through the check that the performance achieved by our approach is in line with the state-of-the-art. We compared the state-of-the-art with our best result, achieved on task \#4, using feature selection on the on-paper features.

Table \ref{tab:compHW} compares our results with those achieved on handwriting data. From the table we can observe that two out of the four approaches compared outperform our approach. However, it must be taken into account that  the compared results have been achieved on a small dataset; this 
%and this makes them statistically few significant and
makes it difficult to perform a fair comparison with our results (achieved on a dataset containing data from 180 participants). Furthermore, it is worth pointing out that, although lower (about 3\%), our performance is close to that of the best-performing handwriting approach (Muller et al. \cite{Mueller2017}). 

Table \ref{tab:compIM} shows the comparison of our results with those achieved on neuroimaging. From the table we can see that our approach  achieved a performance 
%(PERFOMANCES? , MI PIACEREBBE DI ++++) 
similar to those achieved by Abrol et al. \cite{Abrol2020}, whereas outperformed the other ones. However, it must be taken  into account that these results have been achieved using deep learning based approaches or using different sources of data \cite{Abrol2020}. The first provides models that are very difficult to interpret, whereas the second has the problem that some of the data required could not be available. 

%(Ndr MA QUANTO SONO A FAER UNA CHIAVICA LA CONCORRENZA!!!???? ;-)))).

Then we can conclude by stating that this comparison validates our results in the sense mentioned above: the performance of our approach is in line with that of the  state-of-the-art. 
% Anche se i risultati sono stati ottenuti su dati di tipo diverso possiamo sicuramente affermare che questo confronto (valida?) conferma che le performance del nostro approccio sono in linea con lo stao dell'arte. 

\begin{table*}[t]
	\caption{Comparison results with handwriting results.%We considered the performance achieved on Task \#4, on-paper with feature selection for our approach.
}
	\centering
	\begin{tabular}{L{2.2cm}C{1.8cm}cC{1.8cm}c}
		\hline
Year, authors	& Tasks 			 	& Participants  & Classifier 			& Accuracy (\%) \\
		\hline
%		2017,
Garre-Olmo et al. \cite{Garre-Olmo2017} & Sentence writing 	& 40  			& Discriminant Analysis & 82.7 \\
%2017, 
Garre-Olmo et al. \cite{Garre-Olmo2017} & Figure drawing   	& 40  			& Discriminant Analysis & 92.3 \\	
%2017, 
Muller     et al. \cite{Mueller2017} 	  & Figure copying   	& 40  			& Logistic regression   & 92.5 \\
%2017,
Muller     et al. \cite{Mueller2017a}   & Clock drawing test  & 40  			& Logistic regression   & 87.2 \\ \hline
%\multicolumn{2}{l}{
Our approach
%}
& & 174 			& Random Forest      	&  89.0\\ 
		\hline
%		\multicolumn{3}{l}{(*) task 4, on-paper with feature selection.} & & \\		
	\end{tabular}
	\label{tab:compHW}
\end{table*}

\begin{table*}[t]
	\caption{Comparison results with neuroimaging.
%We considered the performance achieved on Task \#4, on-paper with feature selection for our approach.
}
	\centering
\begin{tabular}{L{2.2cm}C{1.8cm}cC{1.8cm}c}
		\hline
Year, authors		& Neuroimaging modality & Participants & Classifier & Acc. (\%) \\
		\hline
%2020, 
Abrol et al. \cite{Abrol2020} & MRI   & 394  & ResNet & 89.3 \\	
%2020, 
Lin et al. \cite{Lin2020} & MRI + PET  & 302 & Lasso & 84.7 \\
%2021, 
Bae et al. \cite{Bae2021}	& MRI & 2490 & CNN & 82.4 \\
%2021, 
Pan et al. \cite{Pan2020} & PET & 479  & CNN & 83.0 \\ \hline
%\multicolumn{2}{l}{
Our approach
%}
& & 174 			& Random Forest      	&  89.0\\ 
\hline
%\multicolumn{2}{l}{Our approach$^*$} & Random Forest & 89.0\\ \hline
%		\multicolumn{2}{l}{(*) task 4, on-paper with feature selection.} & & \\
	\end{tabular}
	\label{tab:compIM}
\end{table*}

\section{Conclusions}% and future work}
\label{sec:concl}
This paper presents the results of a study in which six handwriting tasks have been considered for supporting the diagnosis of AD. These tasks require copying regular words, non-regular words, and non-words. These words represent stimuli with different lexical values (semantics). Our research started from the analysis of the studies conducted on children with different developmental disorders, which demonstrated that the type of word can influence the motor plane needed to write it. We used an innovative approach based on machine learning to investigate how phonology and semantics of words affect the handwriting of people affected. To this aim, we used four well-known and widely-used classifiers and feature selection.  

The experimental results showed that regular words are more discriminant than the other categories and that the in-air movements of people affected by AD are more altered while copying these kinds of words.

The experimental results also demonstrated the effectiveness of  feature selection in improving the performance achieved using the whole set of features extracted, especially for non-regular and non-words.  The feature selection results highlighted the importance of on-paper features in discriminating the handwriting of AD patients while writing non-regular words and non-words. This result confirms that AD harms the brain areas devoted to processing the visual feedback provided by the ink trace and consequently correct the running motor plane already. 
This effect is particularly accentuated for non-regular words and non-words since, in these cases, the recognition of the word and the subsequent recovery of the motor plan requires a greater cognitive effort than for regular words, for which the grapheme-phoneme correspondence allows performing the task with less cognitive effort.

Finally, we compared our results with those achieved  on handwriting data and on brain images to investigate the effectiveness of our machine learning approach in discriminating AD patients. The comparison showed that our results are in line with those of the state-of-the-art, validating the above-discussed findings.

%\section*{conflict of interest}
%No conflict of interest are declared.

%\bibliographystyle{sn-basic}
\bibliography{AIM23}

\begin{thebibliography}{10}
\expandafter\ifx\csname url\endcsname\relax
  \def\url#1{\texttt{#1}}\fi
\expandafter\ifx\csname urlprefix\endcsname\relax\def\urlprefix{URL }\fi
\expandafter\ifx\csname href\endcsname\relax
  \def\href#1#2{#2} \def\path#1{#1}\fi

\bibitem{Shapiro2010}
L.~A. Shapiro, Embodied Cognition, New York: Routledge, 2010.

\bibitem{Impedovo19}
D.~Impedovo, G.~Pirlo, Online Handwriting Analysis for the Assessment of
  Alzheimer’s Disease and Parkinson’s Disease: Overview and Experimental
  Investigation, World Scientific, 19, Ch.~7, pp. 113--128.

\bibitem{Impedovo18}
D.~Impedovo, G.~Pirlo, Dynamic handwriting analysis for the assessment of
  neurodegenerative diseases: a pattern recognition perspective, IEEE Reviews
  in Biomedical Engineering (2018) 1--13.

\bibitem{Nicoll19}
J.~A.~R. Nicoll, G.~R. Buckland, C.~H. Harrison, A.~Page, S.~Harris, S.~Love,
  J.~W. Neal, C.~Holmes, D.~Boche, {Persistent neuropathological effects 14
  years following amyloid-$\beta$ immunization in Alzheimer's disease}, Brain
  142~(7) (2019) 2113--2126.

\bibitem{Drot16}
P.~Drot{\'a}r, J.~Mekyska, I.~Rektorov{\'a}, L.~Masarov{\'a}, Z.~Sm{\'e}kal,
  M.~Fa{\'u}ndez-Zanuy, Evaluation of handwriting kinematics and pressure for
  differential diagnosis of parkinson's disease, Artificial Intelligence in
  Medicine 67 (2016) 39--46.

\bibitem{Cilia18b}
N.~D. Cilia, C.~{De Stefano}, F.~Fontanella, A.~{Scotto di Freca}, An
  experimental protocol to support cognitive impairment diagnosis by using
  handwriting analysis, Procedia Computer Science 141 (2018) 466 -- 471.

\bibitem{Cilia22}
N.~D. Cilia, G.~{De Gregorio}, C.~{De Stefano}, F.~Fontanella, A.~Marcelli,
  A.~Parziale, Diagnosing alzheimer’s disease from on-line handwriting: A
  novel dataset and performance benchmarking, Engineering Applications of
  Artificial Intelligence 111 (2022) 104822.

\bibitem{Cilia21b}
N.~D. Cilia, C.~De~Stefano, F.~Fontanella, A.~S.~D. Freca, Feature selection as
  a tool to support the diagnosis of cognitive impairments through handwriting
  analysis, IEEE Access 9 (2021) 78226--78240.

\bibitem{McRae13}
K.~McRae, M.~N. Jones, Semantic memory., In The Oxford handbook of cognitive
  psychology. Edited by D. Reisberg (2013) 206--216.

\bibitem{Kumar2021}
A.~A. Kumar, Semantic memory: A review of methods, models, and current
  challenges, Psychonomic Bulletin \& Review 28 (2021) 40--80.

\bibitem{Luzzatti03}
C.~Luzzatti, M.~Laiacona, D.~Agazzi, Multiple patterns of writing disorders in
  dementia of the alzheimer-type and their evolution, Neuropsychologia 41~(7)
  (2003) 759--772.

\bibitem{Vessio19}
G.~Vessio, Dynamic handwriting analysis for neurodegenerative disease
  assessment: A literary review, Applied Sciences 9~(21) (2019).

\bibitem{Kolinsky22}
R.~Kolinsky, M.~Tossonian, Phonological and orthographic processing in basic
  literacy adults and dyslexic children, Springer, 2022.

\bibitem{Cilia19c}
N.~D. Cilia, C.~De~Stefano, F.~Fontanella, A.~S. di~Freca, How word choice
  affects cognitive impairment detection by handwriting analysis: A preliminary
  study, in: F.~Cicirelli, A.~Guerrieri, C.~Pizzuti, A.~Socievole, G.~Spezzano,
  A.~Vinci (Eds.), Artificial Life and Evolutionary Computation, Springer
  International Publishing, Cham, 2020, pp. 113--123.

\bibitem{Weekes97}
B.~Weekes, Differential effects of number of letters on word and nonword naming
  latency, The Quarterly Journal of Experimental Psychology 50~(A) (1997)
  439--456.

\bibitem{Barca02}
L.~Barca, C.~Burani, L.~Arduino, Word naming times and psycholinguistic norms
  for italian nouns, Behavior Research Methods, Instruments, \& Computers 34
  (2002) 424--434.

\bibitem{Zoccolotti05}
P.~Zoccolotti, M.~D. Luca, G.~D. Filippo, A.~Judica, D.~Spinelli, Prova di
  lettura di parole e non parole, Roma: IRCCS Fondazione Santa Lucia (2005).

\bibitem{Hasen21}
J.~Hasenäcker, O.~Solaja, D.~Crepaldi, Does morphological structure modulate
  access to embedded word meaning in child readers?, Memory \& Cognition 49~(7)
  (2021) 1334--1347.

\bibitem{Richlan20}
F.~Richlan, The functional neuroanatomy of developmental dyslexia across
  languages and writing systems, Frontiers in Psychology 11 (2020) 155.

\bibitem{Plat93}
H.~Platel, J.~Lambert, F.~Eustache, B.~Cadet, M.~Dary, F.~Viader,
  B.~Lechevalier, Characteristics and evolution of writing impairment in
  alzheimer's disease, Neuropsychologia 31~(11) (1993) 1147--58.

\bibitem{Pestell00}
S.~Pestell, M.~F. Shanks, J.~Warrington, A.~Venneri, Quality of spelling
  breakdown in alzheimer's disease is independent of disease progression,
  Journal of Clinical and Experimental Neuropsychology 22~(5) (2000) 599--612.

\bibitem{Platel93}
H.~Platel, J.~Lambert, F.~Eustache, B.~Cadet, M.~Dary, F.~Viader,
  B.~Lechevalier, Characteristics and evolution of writing impairmant in
  alzheimer's disease, Neuropsychologia 31~(11) (1993) 1147--1158.

\bibitem{Impedovo19b}
D.~Impedovo, G.~Pirlo, Dynamic handwriting analysis for the assessment of
  neurodegenerative diseases: A pattern recognition perspective, IEEE Reviews
  in Biomedical Engineering 12 (2019) 209--220.

\bibitem{RFcit}
L.~Breiman, Random forests, Machine Learning 45~(1) (2001) 5--32.

\bibitem{Kee03}
S.~S. Keerthi, C.-J. Lin, Asymptotic behaviors of support vector machines with
  gaussian kernel, Neural Computation 15~(7) (2003) 1667--1689.

\bibitem{Rumelhart}
D.~E. Rumelhart, G.~E. Hinton, R.~J. Williams, Learning representations by
  back-propagating errors, Nature 323~(9) (1986) 533--536.

\bibitem{Guyon03}
I.~Guyon, A.~Elisseeff, An introduction to variable and feature selection, J.
  Mach. Learn. Res. 3 (2003) 1157--1182.

\bibitem{Ebra20}
M.~A. Ebrahimighahnavieh, S.~Luo, R.~Chiong, Deep learning to detect
  alzheimer's disease from neuroimaging: A systematic literature review,
  Computer Methods and Programs in Biomedicine 187 (2020) 105242.

\bibitem{Goyal2021}
P.~Goyal, R.~Rani, K.~Singh, State-of-the-art machine learning techniques for
  diagnosis of alzheimer’s disease from mr-images: A systematic review,
  Archives of Computational Methods in Engineering (2021) 1--44.

\bibitem{Mueller2017}
S.~M{\"u}ller, O.~Preische, P.~Heymann, U.~Elbing, C.~Laske, Increased
  diagnostic accuracy of digital vs. conventional clock drawing test for
  discrimination of patients in the early course of alzheimer’s disease from
  cognitively healthy individuals, Frontiers in aging neuroscience 9 (2017)
  101.

\bibitem{Abrol2020}
A.~Abrol, M.~Bhattarai, A.~Fedorov, Y.~Du, S.~Plis, V.~Calhoun, A.~D.~N.
  Initiative, et~al., Deep residual learning for neuroimaging: an application
  to predict progression to alzheimer’s disease, Journal of neuroscience
  methods 339 (2020) 108701.

\bibitem{Garre-Olmo2017}
J.~Garre-Olmo, M.~Fa{\'u}ndez-Zanuy, K.~L{\'o}pez-de Ipi{\~n}a,
  L.~Calv{\'o}-Perxas, O.~Turr{\'o}-Garriga, Kinematic and pressure features of
  handwriting and drawing: preliminary results between patients with mild
  cognitive impairment, alzheimer disease and healthy controls, Current
  Alzheimer research 14~(9) (2017) 960--968.

\bibitem{Mueller2017a}
S.~M{\"u}ller, O.~Preische, P.~Heymann, U.~Elbing, C.~Laske, Diagnostic value
  of a tablet-based drawing task for discrimination of patients in the early
  course of alzheimer’s disease from healthy individuals, Journal of
  Alzheimer's Disease 55~(4) (2017) 1463--1469.

\bibitem{Lin2020}
W.~Lin, Q.~Gao, J.~Yuan, Z.~Chen, C.~Feng, W.~Chen, M.~Du, T.~Tong, Predicting
  alzheimer’s disease conversion from mild cognitive impairment using an
  extreme learning machine-based grading method with multimodal data, Frontiers
  in aging neuroscience 12 (2020) 77.

\bibitem{Bae2021}
J.~Bae, J.~Stocks, A.~Heywood, Y.~Jung, L.~Jenkins, V.~Hill, A.~Katsaggelos,
  K.~Popuri, H.~Rosen, M.~F. Beg, et~al., Transfer learning for predicting
  conversion from mild cognitive impairment to dementia of alzheimer's type
  based on a three-dimensional convolutional neural network, Neurobiology of
  aging 99 (2021) 53--64.

\bibitem{Pan2020}
X.~Pan, T.-L. Phan, M.~Adel, C.~Fossati, T.~Gaidon, J.~Wojak, E.~Guedj,
  Multi-view separable pyramid network for ad prediction at mci stage by 18
  f-fdg brain pet imaging, IEEE Transactions on Medical Imaging 40~(1) (2020)
  81--92.

\end{thebibliography}
\end{document}